\RequirePackage{fix-cm}
\documentclass[onecolumn,smallextended,final]{svjour3}
\usepackage[T1]{fontenc}
\usepackage{array}
\usepackage{float}
\usepackage{units}
\usepackage{amsmath}
\usepackage{graphicx}
\usepackage{amssymb}
\usepackage{booktabs}
\usepackage[numbers]{natbib}
\usepackage{epstopdf}

\makeatletter

\newenvironment{lyxlist}[1]
{\begin{list}{} {\settowidth{\labelwidth}{#1}
\setlength{\leftmargin}{\labelwidth}
 \addtolength{\leftmargin}{\labelsep}
 }}
{\end{list}}

\makeatother

\begin{document}

\title{A Review of Feature Selection Methods Based on Mutual Information}
\author{Jorge R. Vergara \and Pablo A. Est\'{e}vez}
\institute{Jorge R. Vergara \at
              Department of Electrical Engineering, Faculty of Physical and Mathematical Sciences, University of Chile, Chile\\
              Tel.: +56-2-29784207\\
              \email{jorgever@ing.uchile.cl}
           \and
           Pablo A. Est\'evez \at
              Department of Electrical Engineering and Advanced Mining Technology Center, Faculty of Physical and Mathematical Sciences, University of Chile, Chile\\
              Tel.: +56-2-29784207\\
              Fax: +56-2-26953881\\
              \email{pestevez@ing.uchile.cl}
}

\date{Received: date / Accepted: date}
\maketitle

\abstract{In this work we present a review of the state of the art of information theoretic feature selection methods.  The concepts of feature relevance, redundance and complementarity (synergy) are clearly defined, as well as Markov blanket. The problem of optimal feature selection is defined. A unifying theoretical framework is described, which can retrofit successful heuristic criteria, indicating the approximations made by each method. A number of open problems in the field are presented.
 }

\section{Introduction}

Feature selection has been widely investigated and used by the machine learning and data mining community. In this context, a feature, also called attribute or variable, represents a property of a process or system than has been measured, or constructed from the original input variables. The goal of feature selection is to select the smallest feature subset given a certain generalization error, or alternatively finding the best feature subset with $k$ features, that yields the minimum generalization error. Additional objectives of feature selection are: (i) improve the generalization performance with respect to the model built using the whole set of features, (ii) provide a more robust generalization and a faster response with unseen data, and (iii) achieve a better and simpler understanding of the process that generates the data \cite{Kohavi97,Guyon03}. We will assume that the feature selection method is used either as a preprocessing step or in conjunction with a learning machine for classification or regression purposes.
Feature selection methods are usually classified in three main groups: wrapper, embedded and filter methods \cite{Guyon03}. Wrappers \cite{Kohavi97} use the induction learning algorithm as part of the function evaluating feature subsets. The performance is usually measured in terms of the classification rate obtained on a testing set, i.e., the classifier is used as a black box for assessing feature subsets. Although these techniques may achieve a good generalization, the computational cost of training the classifier a combinatorial number of times becomes prohibitive for high dimensional datasets. In addition, many classifiers are prone to over-learning and show sensitiveness to initialization. Embedded methods \cite{Lal06}, incorporate knowledge about the specific structure of the class of functions used by a certain learning machine, e.g. bounds on the leave-one-out error of SVMs \cite{Weston00}. Although usually less computationally expensive than wrappers, embedded methods still are much slower than filter approaches, and the features selected are dependent on the learning machine. Filter methods \cite{Duch03} assume complete independence between the learning machine and the data, and therefore use a metric independent of the induction learning algorithm to assess feature subsets. Filter methods are relatively robust against overfitting, but may fail to select the best feature subset for classification or regression.
In the literature, several criteria have been proposed to evaluate single features or feature subsets, among them: inconsistency rate \cite{Huang2003Inconsistency}, inference correlation \cite{Mo2011Inference}, classification error \cite{Estevez1998Clas}, fractal dimension \cite{Mo2012Fractal}, distance measure \cite{Sebban2002Dist,Bins2001Dist}, etc. Mutual information (MI) is a measure of statistical independence, that has two main properties. First, it can measure any kind of relation between random variables, including nonlinear relationships \cite{Cover06}. Second, MI is invariant under transformations in the feature space that are invertible and differentiable, e.g. translations, rotations and any transformation preserving the order of the original elements of the feature vectors \cite{Kullback97,Kullback51}. Many advances in the field have been reported in the last 20 years since the pioneer work of Battiti \cite{Battiti94}. Battiti defined the problem of feature selection as the process of selecting the $k$ most relevant variables from an original feature set of $m$ variables, $k<m$. Battiti proposed the greedy selection of a single feature at a time, as an alternative to evaluate the combinatorial explosion of all feature subsets belonging to the original set. The main assumptions of Battiti\textquoteright s work were the following: (a) features are classified as relevant and redundant; (b) an heuristic functional is used to select features, which allows controlling the tradeoff between relevancy and redundancy; c) a greedy search strategy is used; and d) the selected feature subset is assumed optimal. These four assumptions will be revisited in this work to include recent work on a) new definitions on relevant features and other types of features, b) new information-theoretic functional derived from first principles, c) new search strategies, and d) new definitions of optimal feature subset.
In this work, we present a review of filtering feature selection methods based on mutual information, under a unified theoretical framework. We show the evolution of feature selection methods on the last 20 years, describing advantages and drawbacks. The remainder of this work is organized as follows. In section 2 a background on MI is presented. In section 3, the concepts of relevant, redundant and complementary features are defined. In section 4, the problem of optimal feature selection is defined. In section 5, a unified theoretical framework is presented, which allows us to show the evolution of different MI feature selection methods, as well as their advantages and drawbacks. In section 6, a number of open problems in the field are presented. Finally, in section 7, we present the conclusions of this work.

\section{Background on MI}

\subsection{Notation}

In this work we will use only discrete random variables, because in practice the variables used in most feature selection problems are either discrete by nature or by quantization. Let $F$ be a feature set and $C$ an output vector representing the classes of a real process. Let\textquoteright s assume that $F$ is the realization of a random sampling of an unknown distribution, where $f_{i}$ is the i-th variable of $F$ and $f_{i}(j)$ is the j-th sample of vector $f_{i}$. Likewise, $c_{i}$ is the i-th component of $C$ and $c_{i}(j)$ is the j-th sample of vector $c_{i}$. Uppercase letters denote random sets of variables, and lowercase letters denote individual variables from these sets.

Other notations and terminologies used in this work are the following:
\begin{lyxlist}{00.00.0000}
\item [{$S$}] Subset of current selected variables.
\item [{$f_{i}$}] Candidate feature to be added to or deleted from the subset of selected features $S$.
\item [{$\{f_{i},f_{j}\}$}] Subset composed of the variables $f_{i}$ and $f_{j}$.
\item [{$\neg f_{i}$}] All variables in $F$ except $f_{i}$. $\neg f_{i}=F\setminus f_{i}$.
\item [{$\{f_{i},S\}$}] Subset composed of variable $f_{i}$ and subset $S$.
\item [{$\neg\{f_{i},S\}$}] All variables in $F$ except the subset $\{f_{i},S\}$. $\neg\{f_{i},S\}=F\backslash\{f_{i},S\}$
\item [$p(f_i,C)$] Joint mass probability between variables $f_i$ and $C$.
\item [{$|\cdot|$}] Absolute value / cardinality of a set.
\end{lyxlist}

The sets mentioned above are related as follows: $F=f_{i}\,\cup\,S\,\cup\,\neg\{f_{i},S\}$, $\emptyset=f_{i}\,\cap\, S\,\cap\, \neg\{f_{i},S\}$. The number of samples in $F$ is $n$ and the total number of variables in $F$ is $m$.

\subsection{Basic Definitions}

Entropy, divergence and mutual information are basic concepts defined within information theory \cite{Cover06}. In its origin, information theory was used within the context of communication theory, to find answers about data compression and transmission rate \cite{Shannon48}. Since then, information theory principles have been largely incorporated into machine learning, see for example Principe \cite{Principe}.

\subsubsection{Entropy}
Entropy $(H)$ is a measure of uncertainty of a random variable. The uncertainty is related to the probability of occurrence of an event. Intuitively, high entropy means that each event has about the same probability of occurrence, while low entropy means that each event has a different probability of occurrence. Formally, the entropy of a discrete random variable $x$, with mass probability $p(x(i))=Pr\{x=x(i)\}$, $x(i)\in x$ is defined as:
\begin{equation}
H(x)=-\sum_{i=1}^{n}p(x(i))\log_{2}(p(x(i))).\label{eq:H}
\end{equation}

Entropy is interpreted as the expected value of the negative of the logarithm of mass probability.
Let $x$ and $y$ be two random discrete variables. The joint entropy of $x$ and $y$, with joint mass probability
$p(x(i),y(j))$, is the sum of the uncertainty contained by the two variables. Formally, joint entropy is defined as follows:
\begin{equation}
H(\left\{ x,y\right\} )=-\sum_{i=1}^{n}\sum_{j=1}^{n}p(x(i),y(j))\cdot\log_{2}(p(x(i),y(j))).\label{eq:Hconj}
\end{equation}

The joint entropy has values in the range,
\begin{equation}
\max\left(H(x),H(y)\right)\leq H(\left\{ x,y\right\} )\leq H(x)+H(y).\label{eq:LIM_Hconj}
\end{equation}

The maximum value in inequality (\ref{eq:LIM_Hconj}), happens when $x$ and $y$ are completely independent. The minimum value occurs when $x$ is completely dependent on $y$.
The conditional entropy measures the remaining uncertainty of the random variable $x$ when the value of the random variable $y$ is known. The minimum value of the conditional entropy is zero, and it happens when $x$ is statistically dependent on $y$, i.e., there is no uncertainty in $x$ if we know $y$. The maximum value happens when $x$ and $y$ are statistically independent, i.e., the variable $y$ does not add information to reduce the uncertainty of $x$. Formally, the conditional entropy is defined as:
\begin{equation}
H(x|y)=\sum_{j=1}^{n}p(y(j))\cdot H(x|y=y(j))\label{eq:Hcond}
\end{equation}
where,
\begin{equation}
0<H(x|y)<H(x),\label{eq:LIM_Hcond}
\end{equation}
and $H(x|y=y(j))$ is the entropy of all $x(i)$, which are associated with $y=y(j)$.

Another way of representing the conditional entropy is:
\begin{equation}
H(x|y)=H(\left\{ x,y\right\} )-H(y).
\end{equation}

\subsubsection{Mutual Information}
The mutual information (MI) is a measure of the amount of information that one random variable has about another variable \cite{Cover06}. This definition is useful within the context of feature selection because it gives a way to quantify the relevance of a feature subset with respect to the output vector $C$. Formally, the MI is defined as follows:
\begin{equation}
I(x;y)=\sum_{i=1}^{n}\sum_{j=1}^{n}p(x(i),y(j))\cdot\log\left(\frac{p(x(i),y(j))}{p(x(i))\cdot p(y(j))}\right),
\end{equation}
where MI is zero when $x$ and $y$ are statistically independent, i.e., $p(x(i),y(j))=p(x(i))\cdot p(y(j))$. The MI is related linearly to entropies of the variables through the following equations:
\begin{equation} I(x;y)  = \begin{cases} H(x)-H(x|y)\\ H(y)-H(y|x)\\
H(x)+H(y)-H(x,y).\end{cases}\label{eq:IMenHs}
\end{equation}

Fig.~\ref{fig:Figura1} shows a Venn diagram with the relationships described in (\ref{eq:IMenHs}).

\begin{center}
\begin{figure}[h]
\centering{}\includegraphics[scale=0.4]{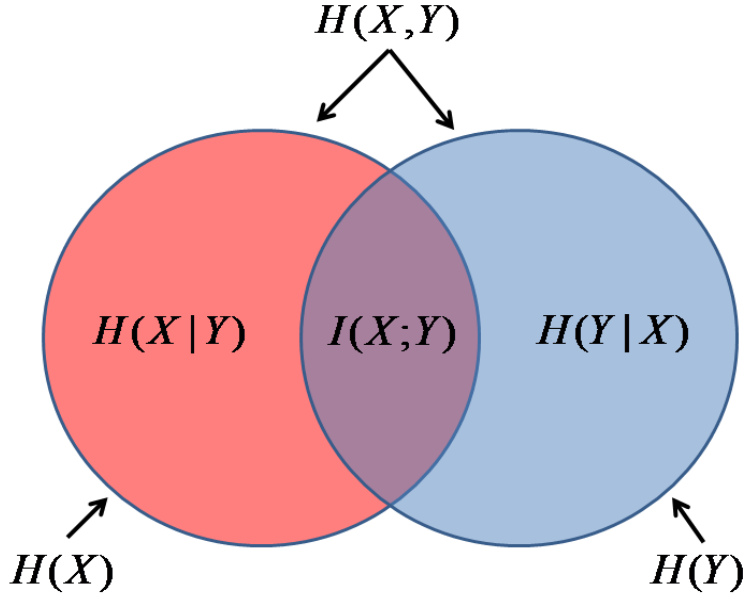}\caption{\label{fig:Figura1}Venn diagram showing the relations between MI and entropies}
\end{figure}
\par\end{center}

Let $z$ be a discrete random variable. Its interaction with the other two variables $\{x,y\}$ can be measured by the conditional MI, which is defined as follows:
\begin{equation}
I(x;y|z)=\sum_{i=1}^{n}p(z(i))I\left(x;y|z=z(i)\right),
\end{equation}
where $I\left(x;y|z=z(i)\right)$ is the MI between $x$ and $y$ in the context of $z=z(i)$. The conditional MI allows measuring the information of two variables in the context of a third one, but it does not measure the information among the three variables. Multi-information is an interesting extension of MI, proposed by McGill \cite{McGill1954}, which allows measuring the interaction among more than two variables. For the case of three variables, the multi-information is defined as follows:
\begin{equation}
I(x;y;z) = \begin{cases}I(\{x,y\};z)-I(x;z)-I(y;z) \\
 I(y;z|x)-I(y;z).\end{cases}\label{eq:INTERACT}
\end{equation}

The multi-information is symmetrical, i.e., $I(x;y;z)=I(x;z;y)=I(z;y;x)$ $=I(y;x;z)=...$ The multi-information has not been widely used in the literature, due to its difficult interpretation, e.g. the multi-information can take negative values, among other reasons. However, there are some interesting papers about the interaction among variables that use this concept \cite{McGill1954,Zhao:2009:SIF:1551582.1551585,Jakulin03,Bell2003}. The multi-information can be understood as the amount of information common to all variables (or set of variables), but that is not present in any subset of these variables. To better understand the concept of multi-information within the context of feature selection, let us consider the following example.

\begin{example}
Let $x_{1},x_{2},x_{3}$ be independent binary random variables. The output of a given system is built through the function $C=x_{1}+\left(x_{2}\oplus
x_{3}\right)$, and $x_{4}=x_{1}$, where $+$ stands for the OR logic function and $\oplus$ represents the XOR logic function.
\begin{center} \begin{tabular}{|c|c|c|c|c|c|} \hline $x_{1}$ & $x_{2}$ & $x_{3}$ & $x_{4}$ &
$x_{2}\oplus x_{3}$ & $C$\tabularnewline \hline \hline 0 & 0 & 0 & 0 & 0 & 0\tabularnewline
\hline 1 & 0 & 0 & 1 & 0 & 1\tabularnewline \hline 0 & 1 & 0 & 0 & 1 & 1\tabularnewline \hline 1
& 1 & 0 & 1 & 1 & 1\tabularnewline \hline 0 & 0 & 1 & 0 & 1 & 1\tabularnewline \hline 1 & 0 & 1 &
1 & 1 & 1\tabularnewline \hline 0 & 1 & 1 & 0 & 0 & 0\tabularnewline \hline 1 & 1 & 1 & 1 & 0 &
1\tabularnewline \hline \end{tabular} \par
\end{center}
\end{example}

Using eq. (\ref{eq:INTERACT}) to measure the multi-information among $x_{2}$, $x_{3}$ and $C$  gives: $I(x_{2};x_{3};C)=I(\{x_{2},x_{3}\};C)-I(x_{2};C)-I(x_{3};C)$. Notice that the relevance of single features $x_{2}$ and $x_{3}$ with respect to $C$ is null, since $I(x_{2};C)=I(x_{3};C)=0$, but the joint information of $\{x_{2},x_{3}\}$ with respect to $C$ is greater than zero, $I(\{x_{2},x_{3}\};C)>0$. In this case, $x_{2}$ and $x_{3}$ interact positively to predict $C$, and this yields a positive value of the multi-information among these variables. The multi-information among the variables $x_{1}$, $x_{4}$ and $C$ is given by: $I(x_{1};x_{4};C)=I(\{x_{1},x_{4}\};C)-I(x_{1};C)-I(x_{4};C)$. The relevance of individual features $x_{1}$ and $x_{4}$ is the same, i.e., $I(x_{1};C)=I(x_{4};C)>0$. In this case the joint information provided by $x_{1}$ and $x_{4}$ with respect to $C$ is the same as that of each variable acting separately, i.e., $I(\{x_{1},x_{4}\};C)=I(x_{1};C)=I(x_{4};C)$. This yields a negative value of the multi-information among these variables. We can deduce that the interaction between $x_{1}$ and $x_{4}$ does not provide any new information about $C$. Let us consider now the multi-information among $x_{1}$, $x_{2}$ and $C$, which is zero: $I(x_{1};x_{2};C)=I(\{x_{1},x_{2}\};C)-I(x_{1};C)-I(x_{2};C)=0$. Since feature $x_{2}$ only provides information about $C$ when interacting with $x_{3}$, then $I(\{x_{1},x_{2}\};C)=I(x_{1};C)$. In this case, features $x_{1}$ and $x_{2}$ do not interact in the knowledge of $C$.

From the viewpoint of feature selection, the value of the multi-information (positive, negative or zero) gives rich information about the kind of interaction there is among the variables. Let us consider the case where we have a set of already selected features $S$ and a candidate feature $f_{i}$, and we measure the multi-information of these variables with the class variable $C$, $I(f_{i};S;C)=I(S;C|f_{i})-I(S;C)$. When the multi-information is positive, it means that feature $f_i$ and $S$ are complementary. On the other hand, when the multi-information is negative, it means that by adding $f_{i}$ we are diminishing the dependence between $S$ and $C$, because $f_{i}$ and $S$ are redundant. Finally, when the multi-information is zero, it means that $f_{i}$ is irrelevant with respect to the dependency between $S$ and $C$.

The mutual information between a set of $m$ features and the class variable $C$ can be expressed compactly in terms of multi-information as follows:
\begin{equation}
I(\left\{ x_{1},x_{2},...,x_{m}\right\} ;C)=\sum_{k=1}^{m}\sum_{\begin{array}{c}\forall S\subseteq\{x_{1},...,x_{m}\}\\ |S|=k\end{array}}I([S\cup C]),\label{eq:Multiple}
\end{equation}
where $I([S\cup C])=I(s_{1};s_{2};\cdots;s_{k};C).$ Note that the sum on the right side of eq. (\ref{eq:Multiple}), is taken over all subsets $S$ of size $k$ drawn from the set $\{x_{1},...,x_{m}\}$.

\section{Relevance, Redundancy and Complementarity}

The filter approach to feature selection is based on the idea of relevance, which we will explore in more detail in this section. Basically the problem is to find the feature subset of minimum cardinality that preserves the information contained in the whole set of features with respect to $C$. This problem is usually solved by finding the relevant features and discarding redundant and irrelevant features. In this section, we review the different definitions of relevance, redundancy and complementarity found in the literature.

\subsection{Relevance}

Intuitively, a given feature is relevant when either individually or together with other variables, it provides information about $C$. In the literature there are many definitions of relevance, including different levels of relevance \cite{Bell00,Battiti94,Guyon03,Kohavi97,Yu04,Peng05,Almuallim92,Almuallim91,Brown2012,Davies94}. Kohavi and John \cite{Kohavi97} used a probabilistic framework to define three levels of relevance: strongly relevant, weakly relevant, and irrelevant features, as shown in Table \ref{tab:relevance}. Strongly relevant features provide unique information about $C$, i.e., they cannot be replaced by other features. Weakly relevant features provide information about $C$, but they can be replaced by other features without losing information about $C$. Irrelevant features do not provide information about $C$, and they can be discarded without losing information. A drawback of the probabilistic approach is the need of testing the conditional independence for all possible feature subsets, and estimating the probability density functions (pdfs) \cite{Raudy1991}.

An alternative definition of relevance is given under the framework of mutual information \cite{Somol04,Bell00,Kojadinovic05,Koller96,Yu04,Kwak02b,Fleuret04,Tishby99}. An advantage of this approach is that there are several good methods for estimating MI. The last column of Table \ref{tab:relevance} shows how the three levels of individual relevance are defined in terms of MI.
\begin{table}[H]
\caption{\label{tab:relevance}Levels of relevance for candidate feature $f_{i}$, according to probabilistic framework \cite{Kohavi97} and mutual information framework \cite{Meyer08}}
\centering{}\begin{tabular}{>{\centering}m{2cm}|c|>{\centering}m{3.7cm}|>{\centering}m{2.8cm}}
\hline Relevance Level & Condition  & Probabilistic Approach & Mutual Information Approach\tabularnewline \hline \hline Strongly\\
 Relevant & $\nexists$ & $p(C|f_{i},\neg f_{i})\neq p(C|\neg f_{i})$ & $I(f_{i};C|\neg
 f_{i})>0$\tabularnewline
\hline Weakly\\
 Relevant & $\exists\, S\subset\neg f_{i}$ & $p(C|f_{i},\neg f_{i})=p(C|\neg f_{i})$\\
 {$\,\,\,\,\,\,\,\,\,\,\,\,\,\,\wedge\,\,\,\,\,\,\,\,\,\,\,\,\,\,$}
$p(C|f_{i},S)\neq p(C|S)$ & $I(f_{i};C|\neg f_{i})=0$\\
 {$\,\,\,\,\,\,\,\,\wedge\,\,\,\,\,\,\,\,$
$I(f_{i};C|S)>0$}\tabularnewline \hline Irrelevant & $\forall\, S\subseteq\neg f_{i}$ &
$p(C|f_{i},S)=p(C|S)$ & $I(f_{i};C|S)=0$\tabularnewline \end{tabular}
\end{table}

The definitions shown in Table \ref{tab:relevance} give rise to several drawbacks, which are summarized as follows:
\begin{enumerate}
\item To classify a given feature $f_{i}$, as irrelevant, it is necessary to assess all possible subsets $S$ of $\neg f_{i}$. Therefore this procedure is subject to the curse of dimensionality \cite{Bellman61,Trunk1979}.
\item The definition of strongly relevant features is too restrictive. If two features provides information about the class but are redundant, then both features will be discarded by this criterion. For example, let $\{x_{1},x_{2},x_{3}\}$ be a set of 3 variables, where $x_{1}=x_{2}$, and $x_{3}$ is noise, and the output class is defined as $C=x_{1}$. Following the strong relevance criterion we have $I(x_{1};C|\{x_{2},x_{3}\})=$$I(x_{2};C|\{x_{1},x_{3}\})=$$I(x_{3};C|\{x_{1},x_{2}\})=0$.
\item The definition of weak relevance is not enough for deciding whether to discard a feature from the optimal feature set. It is necessary to discriminate between redundant and non-redundant features.
\end{enumerate}

\subsection{Redundancy}

Yu and Liu \cite{Yu04} proposed a finer classification of features into weakly relevant but redundant and weakly relevant but non-redundant. Moreover, the authors defined the set of optimal features as the one composed by strongly relevant features and weakly relevant but non-redundant features. The concept of redundancy is associated with the level of dependency among two or more features. In principle we can measure the dependency of a given feature $f_{i}$ with respect to a feature subset $S\subseteq\text{\ensuremath{\neg}}f_{i}$, by simply using the MI, $I(f_{i};S)$.
This information theoretic measure of redundancy satisfies the following properties: it is symmetric, non-linear, non-negative, and does not diminish when adding new features \cite{Meyer08}.
However, using this measure it is not possible to determine concretely with which features of $S$ is $f_{i}$ redundant. This calls for more elaborated criteria of redundancy, such as the Markov blanket \cite{Koller96,Yu04}, and total correlation \cite{Watanabe1960}. The Markov blanket is a strong condition for conditional independence, and is defined as follows.
\begin{definition}[Markov blanket]
Given a feature $f_{i}$, the subset $M\subseteq\neg f_{i}$ is a Markov blanket
of $f_{i}$ iff \cite{Koller96,Yu04}:
\begin{equation}
p(\{F\backslash\{f_{i}\,,M\},C\}\,\text{|\,}\{f_{i}\,,M\})  =  p(\{F\backslash\{f_{i}\,,M\},C\}\,\text{|\,}M).\label{eq:MB_prob}
\end{equation}
\end{definition}

This condition requires that M subsumes all the information that $f_{i}$ has about $C$, but also about all other features $\{F\backslash\{f_{i}\,,M\}\}$. It can be proved that strongly relevant features do not have a Markov blanket \cite{Yu04}.

The Markov blanket condition given by Eq. (\ref{eq:MB_prob}) can be rewritten in the context of information theory as follows \cite{Meyer08}:
\begin{equation}
I(f_{i};\{C,\neg {f_{i},M}\}\text{|\,}M)=0.\label{MBMI}
\end{equation}

An alternative measure of redundancy is the total correlation or multivariate correlation \cite{Watanabe1960}. Given a set of features $F=\{f_{1},...,f_{m}\}$, the total correlation is defined as follows:
\begin{equation}
C(f_{1};...;f_{m})=\sum_{i=1}^{m}H(f_{i})-H(f_{1},...,f_{m}).\label{eq:CORRT1}
\end{equation}

Total correlation measures the common information (redundancy) among all the variables in $F$. If we want to measure the redundancy between a given variable $f_{i}$ and any feature subset $S\subseteq\neg f_{i}$, then we can use the total correlation as:
\begin{equation}
C(f_{i};S)=H(f_{i})+H(S)-H(f_{i},S),
\end{equation}
however this corresponds to the classic definition of MI, i.e., $C(f_{i};S)=I(f_{i};S)$.

\subsection{Complementarity}

The concept of complementarity has been re-discovered several times \cite{Meyer08,Bonev08,Brown2012,Vidal2003,Cheng2011}. Recently, it has become more relevant because of the development of more efficient techniques to estimate MI in high-dimensional spaces \cite{Kraskov2004,Hero99}. Complementarity, also known as synergy, measures the degree of interaction between an individual feature $f_{i}$ and feature subset $S$ given $C$, through the following expression $(I(f_{i};S|C))$. To illustrate the concept of complementarity, we will start expanding the multi-information among $f_{i}$, $C$ and $S$. Decomposing the multi-information in its three possible expressions we have:
\begin{equation}
I(f_{i};S;C)=
\begin{cases}
I(f_{i};S|C)-I(f_{i};S)\\
I(f_{i};C|S)-I(f_{i};C)\\
I(S;C|f_{i})-I(S;C).
\end{cases}\label{eq:INTERATC1}
\end{equation}

According to eq. (\ref{eq:INTERATC1}), the first row shows that the multi-information can be expressed as the difference between complementarity $(I(f_{i};S|C))$ and redundancy $(I(f_{i};S))$. A positive value of the multi-information entails a dominance of complementarity over redundancy. Analyzing the second row of eq. (\ref{eq:INTERATC1}), we observe that this expression becomes positive when the information that $f_{i}$ has about $C$ is greater when it interacts with subset $S$ with respect to the case when it does not. This effect is called complementarity. The third row of eq. (\ref{eq:INTERATC1}), gives us another viewpoint of the complementarity effect. The multi-information is positive when the information that $S$ has about $C$ is greater when it interacts with feature $f_{i}$ compared to the case when it does not interact. Assuming that the complementarity effect is dominant over redundancy, Fig. \ref{fig:ResRel} illustrates a Venn diagram with the relationships among complementarity, redundancy and relevancy.
\begin{center}
\begin{figure}[h]
\centering{}\includegraphics[scale=0.4]{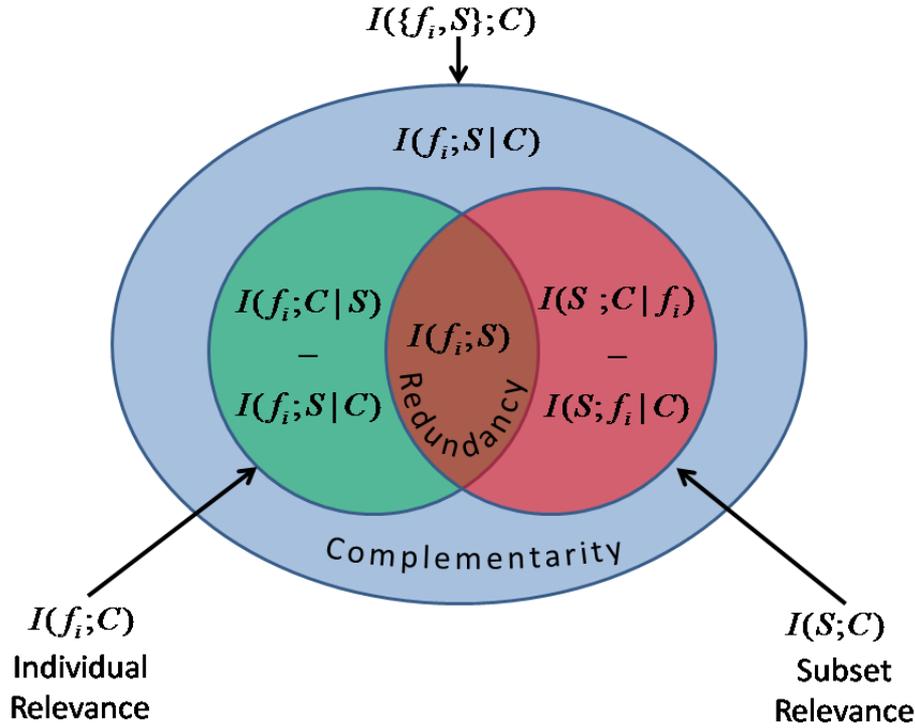}
\caption{\label{fig:ResRel} Venn diagram showing the relationships among complementarity, redundancy and relevancy, assuming that the multi-information among $f_{i}$, $S$ and $C$ is positive.}
\end{figure}
\par\end{center}

\section{Optimal Feature Subset}

In this section we review the different definitions of the optimal feature subset, $S_{opt}$, given in the literature, as well as the search strategies used for obtaining this optimal set. According to \cite{Tsamardinos03c}, in practice the feature selection problem must include a classifier or an ensemble of classifiers, and a performance metric. The optimal feature subset is defined as the one that maximizes the performance metric having minimum cardinality. However, filter methods are independent of both the learning machine and the performance metric. Any filter method corresponds to a definition of relevance that employs only the data distribution \cite{Tsamardinos03c}. Yu and Liu \cite{Yu04} defined the optimal feature set as composed of all strongly relevant features and the weakly relevant but not redundant features. In this section we review the definitions of the optimal feature subset from the viewpoint of filter methods, in particular MI feature selection methods. The key notion is conditional independence, which allows defining the sufficient feature subset as follows \cite{Bell00,GuyonFE}:
\begin{definition}
$S\subseteq F$ is a sufficient feature subset iff
\begin{equation}
p(C|F)=p(C|S).\label{eq:SFS}
\end{equation}
\end{definition}

This definition implies that $C$ and $\neg S$ are conditionally independent, i.e., $\neg S$ provides no additional information about $C$ in the context of $S$. However, we still need a search strategy to select the feature subset $S$, and an exhaustive search using this criterion is impractical due to the curse of dimensionality.

In probability the measure of sufficient feature subset can be expressed as the expected value over $p(F)$ of the Kullback-Leibler divergence between $p(C|F)$ and $p(C|S)$ \cite{Koller96}. According to Guyon \textit{et al.} \cite{GuyonFE}, this can be expressed in terms of MI as follows:
\begin{equation}
DMI(S)=I(F;C)-I(S;C).\label{eq:SFS}
\end{equation}

Guyon \textit{et al.} \cite{GuyonFE} proposed solving the following optimization problem:
\begin{equation}
\min_{S\subseteq F}|S|+\lambda\cdot DMI(S),\label{eq:OSF}
\end{equation}
where $\lambda>0$ represents the Lagrange multiplier. If $S$ is a sufficient feature subset, then $DMI(S)=0$, and eq. (\ref{eq:OSF}) is reduced to $\min_{S\subseteq F}|S|$. Since $I(F;C)$ is constant, eq. (\ref{eq:OSF}) is equivalent to:
\begin{equation}
\min_{S\subseteq F}|S|-\lambda\cdot I(S;C).\label{eq:OFS_MI}
\end{equation}

The feature selection problem corresponds to finding the smallest feature subset that maximizes $I(S;C)$. Since the term $\min_{S\subseteq F}|S|$ is discrete, the optimization of (\ref{eq:OFS_MI}) is difficult. Tishby \textit{et al.} \cite{Tishby99} proposed replacing the term $\min_{S\subseteq F}|S|$ with $I(F;S)$.

An alternative approach to optimal feature subset selection is using the concept of the Markov blanket (MB). Remember that the Markov blanket, $M$, of a target variable $C$, is the smallest subset of $F$ such that $C$ is independent of the rest of the variables $F\backslash M$. Koller and Sahami \cite{Koller96} proposed using MBs as the basis for feature elimination. They proved that features eliminated sequentially based on this criterion remain unnecessary. However, the time needed for inducing an MB grows exponentially with the size of this set, when considering full dependencies. Therefore most MB algorithms implement approximations based on heuristics, e.g. finding the set of $k$ features that are strongly correlated with a given feature \cite{Koller96}. Fast MB discovery algorithms have been developed for the case of distributions that are faithful to a Bayesian Network \cite{Tsamardinos03c,Tsamardinos03b}. However, these algorithms require that the optimal feature subset does not contain multivariate associations among variables, which are individually irrelevant but become relevant in the context of others \cite{BrownyTsamardinos}. In practice, this means for example that current MB discovery algorithms cannot solve Example 1 due to the XOR function.

An important caveat is that both feature selection approaches, sufficient feature subset and MBs, are based on estimating the probability distribution of $C$ given the data. Estimating posterior probabilities is a harder problem than classification, e.g. in using a $0\backslash 1$-loss function only the most probable classification is needed. Therefore, this effect may render some features contained in sufficient feature subset or in the MB of $C$ unnecessary \cite{Tsamardinos03c,Torkkola,GuyonFE}.

\subsection{Relation between MI and Bayes error classification}

There are some interesting results relating the MI between a random discrete variable $f$ and a random discrete target variable $C$, with the minimum error obtained by maximum a posteriori classifier (Bayer classification error) \cite{Hellman70,Cover06,Feder94}. The Bayes error is bounded above and below according to the following expression:
\begin{equation}
1-\frac{I(f;C)+\log(2)}{\log(|C|)}\leq e_{bayes}(f)\leqslant\frac{1}{2}\left(H(C)-I(f;C)\right).\label{eq:IM_lim_inf}
\end{equation}
Interestingly, Eq. (\ref{eq:IM_lim_inf}) shows that both limits are minimized when the MI, $I(f;C)$, is maximized.

\subsection{Search strategies}
\label{search}
According to Guyon \textit{et al.} \cite{GuyonFE}, a feature selection method has three components: 1) Evaluation criterion definition, e.g. relevance for filter methods, 2) evaluation criterion estimation, e.g. sufficient feature selection or MB for filter methods, and 3) search strategies for feature subset generation. In this section, we briefly review the main search strategies used by MI feature selection methods. Given a feature set $F$ of cardinality $m$, there are $2^{m}$ possible subsets, therefore an exhaustive search is impractical for high-dimensional datasets. 

There are two basic search strategies: optimal methods and sub-optimal methods \cite{Webb02}. Optimal search strategies include exhaustive search and accelerated methods based on the monotonic property of a feature selection criterion, such as branch and bound. But optimal methods are impractical for high-dimensional datasets, therefore sub-optimal strategies must be used.

Most popular search methods are sequential forward selection (SFS) \cite{Whitney1971} and sequential backward elimination (SBE) \cite{Marill193}. Sequential forward selection is a bottom-up search, which starts with an empty set, and adds new features one at a time. Formally, it adds the candidate feature $f_i$ that maximizes $I(S;C)$ to the subset of selected features $S$, i.e.,
\begin{equation}
S=S\cup\{\underset{f_{i}\in F\backslash S}{\arg\,\max}(I(\{S,f_{i}\};C))\}.
\end{equation}

Sequential backward elimination is a top-down approach, which starts with the whole set of features, and deletes one feature at a time. Formally, it starts with $S=F$, and proceeds deleting the less informative features one at a time, i.e,
\begin{equation}
S=S\backslash\{\underset{f_{i}\in S}{\arg\,\min}(I(\{S\backslash f_{i}\};C)\}.
\end{equation}
Usually backward elimination is computationally more expensive than forward selection, e.g. when searching for a small subset of features. However, backward elimination can usually find better feature subsets, because most forward selection methods do not take into account the relevance of variables in the context of features not yet included in the subset of selected features \cite{Guyon03}. Both kinds of searching methods suffer from the nested effect, meaning that in forward selection a variable cannot be deleted from the feature set once it has been added, and in backward selection a variable cannot be reincorporated once it has been deleted. Instead of adding a single feature at a time, some generalized forward selection variants add several features, to take into account the statistical relationship between variables \cite{Webb02}. Likewise, the generalized backward elimination deletes several variables at a time. An enhancement may be obtained by combining forward and backward selection, avoiding the nested effect. The strategy ``plus-l-take-away-r'' \cite{Stearns1976} adds to $S$ $l$ features and then removes the worst $r$ features if $l>r$, or deletes $r$ features and then adds $l$ features if $r<l$.

\section{A Unified Framework for Mutual Information Feature Selection}
\label{UF}

Many MI feature selection methods have been proposed in the last 20 years. Most methods define heuristic functionals to assess feature subsets combining definitions of relevant and redundant features. Brown~\textit{et al.} \cite{Brown2012} proposed a unifying framework for information theoretic feature selection methods. The authors posed the feature selection problem as a conditional likelihood of the class labels, given features. Under the filter assumption \cite{Brown2012}, conditional likelihood is equivalent to conditional mutual information (CMI), i.e., the feature selection problem can be posed as follows:
\begin{eqnarray}
 &  & \underset{S\subseteq F}{\min}|S| \label{eq:Brown}\\
 &  & subject\, to:\,\,\underset{S\subseteq F}{\min}I(\neg S;C|S).\nonumber
\end{eqnarray}
This corresponds to the smallest feature subset such that the CMI is minimal.
Starting from this objective function, the authors used MI properties to deduce some common heuristic criteria used for MI feature selection. Several criteria can be unified under the proposed framework. In particular, they showed that common heuristics based on linear combinations of information terms, such as Battiti\textquoteright s MIFS \cite{Battiti94}, conditional infomax feature extraction (CIFE) \cite{Lin2006,Guo09}, minimum-redundancy maximum relevance (mRMR) \cite{Peng05}, and joint mutual information (JMI) \cite{YangMoody99}, are all low-order approximations to the conditional likelihood optimization problem. However, the unifying framework proposed by Brown \textit{et al.} \cite{Brown2012} fell short of deriving (explaining) non-linear criteria using min or max operators such as Conditional Mutual Information Maximization (CMIM) \cite{Fleuret04}, Informative Fragments \cite{Vidal2003}, and ICAP \cite{PHD_Jakulin2005}.

Let us start with the assumption that $I(F;C)$ measures all the information about the target variable contained in the set of features. This assumption is based on the additivity property of MI \cite{Cover06,Kojadinovic05}, which states that the information about a given system is maximal when all features ($F$) are used to estimate the target variable ($C$). Using the chain rule, $I(F;C)$ can be decomposed as follows:
\begin{equation}
I(F;C)=I(S;C)+I(\neg S;C|S).\label{eq:OPTJ}
\end{equation}

As $I(F;C)$ is constant, maximizing $I(S;C)$ is equivalent to minimizing $I(\neg S;C|S)$. Many MI feature selection methods maximize the first term on the right side of (\ref{eq:OPTJ}). This is known as the criterion of maximal dependency (MD) \cite{Peng05}. On the other hand, other criteria are based on the idea of minimizing the CMI, i.e. the second term on the right hand side of eq. ~(\ref{eq:OPTJ}).

In the following we describe the approach of Brown \textit{et al.} \cite{Brown2012} for deriving sequential forward selection and sequential backward elimination algorithms, which are based on minimizing the CMI. For the convenience of the reader, we present the equivalent procedure in parallel when maximizing dependency (MD). In practice, a search strategy is needed to find the best feature subset. As we saw in section ~\ref{search}, the most popular methods are sequential forward selection and sequential backward elimination. Before proceeding we need to define some notation.

\begin{lyxlist}{00.00.0000}
\item [{$S^{t}$}] Subset of selected variables at time t.
\item [{$f_{i}$}] Candidate feature to be added to or eliminated from feature subset $S^{t}$ at time $t$.
\begin{lyxlist}{00.00.0000}
\item [{$f_{i}=\underset{f_{i}\in\neg S^{t}}{\arg\,\max}\,\, I(f_{i};C|S^{t})$}] in forward selection.
\item [{$f_{i}=\underset{f_{i}\in S^{t}}{\arg\,\min}\,\, I(f_{i};C|S^{t}\backslash f_{i})$}] in backward elimination.
\end{lyxlist}
\item [{$s_{j}$}] A given feature in $S^{t}$.
\item [{$\neg s_{j}$}] The complement set of feature $s_j$ with set $S^{t}$, i.e., $\neg s_{j}=S^{t}\backslash s_{j}$
\item [{$S^{t+1}$}] Subset of selected variables at time t+1.
\begin{lyxlist}{00.0000000.0000}
\item [{$S^{t+1}\leftarrow\{S^{t},f_{i}\}$}] in forward selection.
\item [{$S^{t+1}\leftarrow S^{t}\backslash f_{i}$}] in backward elimination.
\end{lyxlist}
\item [{$\neg S^{t+1}$}] Complement of feature subset $S^{t+1}$, i.e. $F=\{S^{t+1},\neg S^{t+1}\}$.
\begin{lyxlist}{00.0000000.0000}
\item [{$\neg S^{t+1}\leftarrow\{\neg S^{t}\backslash f_{i}\}$}] in forward selection.
\item [{$\neg S^{t+1}\leftarrow\{\neg S^{t},f_{i}\}$}] in backward elimination.
\end{lyxlist}
\end{lyxlist}

Table~\ref{tab:SFS} shows that for the case of sequential forward selection, we achieve the same result when using the MD or CMI approach: the SFS algorithm consists of maximizing $I(f_i;C|S^{t})$. Analogously, Table~\ref{tab:BFS} shows that for the case of sequential backward elimination, again we achieve the same result when using MD or CMI approaches: the SBE algorithm consists of minimizing $I(f_i;C|S^{t}\backslash f_{i})$.

\begin{table}
\caption{ \label{tab:SFS}Parallel between MD and CMI approaches for sequential forward selection}
\centering{}%
\begin{minipage}[t]{1\columnwidth}%
\begin{center}
\begin{tabular}{cc}
\toprule
\textbf{MD} & \textbf{CMI}\tabularnewline
\toprule
\midrule
$\underset{f_{i}\in\neg S^{t}}{\max}I(S^{t+1};C)$= & $\underset{f_{i}\in\neg S^{t}}{\min}I(\neg S^{t+1};C|S^{t+1})$ \tabularnewline
$\underset{f_{i}\in\neg S^{t}}{\max}I(\{S^{t},f_{i}\};C)$= & $\underset{f_{i}\in\neg S^{t}}{\min}I(\neg S^{t}\backslash f_{i};C|\{S^{t},f_{i}\})$\tabularnewline
$\underset{f_{i}\in\neg
S^{t}}{\max}I(S^{t};C)$\footnote{This term is independent of $f_{i}$.}
$+\underset{f_{i}\in\neg S^{t}}{\max}I(f_{i};C|S^{t})$ &
$\underset{f_{i}\in\neg S^{t}}{\min}I(\neg S^{t};C|S^{t})$\footnote{This
term has the same value $\forall f_{i}$.}$+\underset{f_{i}\in\neg
S^{t}}{\min}\left(-I(f_{i};C|S^{t})\right)$\tabularnewline
$\Downarrow$ & $\Downarrow$\tabularnewline
$\underset{f_{i}\in\neg S^{t}}{\max}I(f_{i};C|S^{t})$ & $\underset{f_{i}\in\neg S^{t}}{\max}I(f_{i};C|S^{t})$\tabularnewline
\bottomrule
\end{tabular}
\par\end{center}%
\end{minipage}
\end{table}

\begin{table}
\caption{\label{tab:BFS}Parallel between MD and CMI approaches for sequential backward elimination}
\centering{}%
\begin{minipage}[t]{1\columnwidth}%
\begin{center}
\begin{tabular}{cc}
\toprule
\textbf{MD} & \textbf{CMI}\tabularnewline
\toprule
\midrule
$\underset{f_{i}\in S^{t}}{\max}I(S^{t+1};C)$= & $\underset{f_{i}\in
S^{t}}{\min}I(\neg S^{t+1};C|S^{t+1})$ \tabularnewline
$\underset{f_{i}\in S^{t}}{\max}I(S^{t}\backslash f;C)$= & $\underset{f_{i}\in S^{t}}{\min}I(\{\neg S^{t},f_{i}\}\backslash f_{i};C|S^{t}\backslash f_{i})$\tabularnewline
$\underset{f_{i}\in S^{t}}{\max}I(S^{t};C)$ \footnote{This term is independent of $f_{i}$.}$+\underset{f_{i}\in S^{t}}{\max}\left(-I(f_{i};C|S^{t}\backslash f_{i})\right)$ & $\underset{f_{i}\in S^{t}}{\min}I(\neg S^{t};C|S^{t})$ \footnote{This term has the same value $\forall f_{i}$.}$+\underset{f_{i}\in S^{t}}{\min}\left(I(f_{i};C|S^{t}\backslash f_{i})\right)$\tabularnewline
$\Downarrow$ & $\Downarrow$\tabularnewline
$\underset{f_{i}\in S^{t}}{\min}I(f_{i};C|S^{t}\backslash f_{i})$ & $\underset{f_{i}\in S^{t}}{\min}I(f_{i};C|S^{t}\backslash f_{i})$\tabularnewline
\bottomrule
\end{tabular}
\par\end{center}%
\end{minipage}
\end{table}

For space limitations, we will develop here only the case of forward feature selection, but the procedure is analogous for the case of backward feature elimination. The expression {$I(f_{i};C|S^{t})$} can be expanded as follows \cite{Cheng2011}:
\begin{equation}
I(f_{i};C|S^{t}) =I(f_{i};C)-I(f_{i};S^{t})+I(f_{i};S^{t}|C).\label{eq:ed1}
\end{equation}

The first term on the right hand side of (\ref{eq:ed1}) measures the individual relevance of the candidate feature $f_i$ with respect to output $C$; the second term measures the redundance of the candidate feature with the feature subset of previously selected features $S^{t}$; and the third term measures the complementarity between $S^{t}$ and $f_{i}$ in the context of $C$. However, from the practical point of view, eq. (\ref{eq:ed1}) presents the difficulty of estimating MI in high-dimensional spaces, due to the presence of the set $S^{t}$ in the second and third terms.

In what follows, we take a detour from the derivation of Brown \textit{et al.} \cite{Brown2012}, using our own alternative approach. To avoid the previously mentioned problem, $I(f_{i};S^{t})$ with $|S^{t}|=p$ can be calculated by averaging all expansions over every single feature in $S$, by using the chain rule as follows:
\begin{align}
I(f_{i};S^{t}) & =\hphantom{lllllllllll}I(f_{i};s_{1})+\hphantom{lllllllllll}I(f_{i};\neg s_{1}|s_{1})\nonumber \\
I(f_{i};S^{t}) & =\hphantom{lllllllllll}I(f_{i};s_{2})+\hphantom{lllllllllll}I(f_{i};\neg s_{2}|s_{2})\nonumber \\
\vdots\phantom{ssss} & =\hphantom{lllllllllll}\phantom{sss}\vdots\hphantom{ssssssssssss\hphantom{lllllllllll}}\vdots\nonumber \\
I(f_{i};S^{t}) & =\hphantom{lllllllllll}I(f_{i};s_{p})+\hphantom{lllllllllll}I(f_{i};\neg s_{p}|s_{p})\nonumber \\
\overline{\hphantom{zzzzzzzzz}} & \hphantom{ss}\overline{\hphantom{sssssssssssssssssssssssssssssssssssss}}\nonumber \\
I(f_{i};S^{t}) & ={\displaystyle \frac{1}{|S^{t}|}\sum_{s_{j}\in S^{t}}I(f_{i};s_{j})}+\frac{1}{|S^{t}|}\sum_{s_{j}\in S^{t}}I(f_{i};\neg s_{j}|s_{j}).\label{eq:REL1_1}
\end{align}
Analogously, we can obtain the following expansion for the conditional mutual information, $I(f_{i};S^{t}|C)$:
\begin{equation}
I(f_{i};S^{t}|C)={\displaystyle \frac{1}{|S^{t}|}\sum_{s_{j}\in S^{t}}I(f_{i};s_{j}|C)}+\frac{1}{|S^{t}|}\sum_{s_{j}\in S^{t}}I(f_{i};\neg s_{j}|\{C,s_{j}\}).\label{eq:REL1_2}
\end{equation}
Substituting (\ref{eq:REL1_1}) and (\ref{eq:REL1_2}) into eq. (\ref{eq:ed1}) yields:
\begin{align}
I(f_{i};C|S^{t}) & =I(f_{i};C)-\left(\frac{1}{|S^{t}|}{\displaystyle \sum_{s_{j}\in S^{t}}I(f_{i};s_{j})}+\frac{1}{|S^{t}|}\sum_{s_{j}\in S^{t}}I(f_{i};\neg s_{j}|s_{j})\right)\nonumber \\
 & \hphantom{rrrrrrrr}+\left(\frac{1}{|S|}{\displaystyle \sum_{s_{j}\in S}I(f_{i};s_{j}|C)}+\frac{1}{|S|}\sum_{s_{j}\in S}I(f_{i};\neg s_{j}|\{C,s_{j}\})\right).\label{eq:REL_APROX_1}
\end{align}
Eq. (\ref{eq:REL_APROX_1}), can be approximated by considering assumptions of lower-order dependencies between features \cite{Balagani2010}. Features $s_j \in S^{t}$ are assumed to have only one-to-one dependencies with $f_i$ or $C$. Formally, assuming statistical independence:
\begin{eqnarray}
p(f_{i}|S^{t}) & = & \prod_{s_{j}\in S^{t}}p(f_{i}|s_{j})\nonumber \\
p(f_{i}|\{S^{t},C\}) & = & \prod_{s_{j}\in S^{t}}p(f_{i}|\{s_{j},C\}),
\end{eqnarray}
we obtain the following low-order approximation:
\begin{equation}
I(f_{i};C|S^{t})\approx I(f_{i};C)-\frac{1}{|S^{t}|}{\displaystyle \sum_{s_{j}\in S^{t}}I(f_{i};s_{j})}+\frac{1}{|S^{t}|}{\displaystyle \sum_{s_{j}\in S^{t}}I(f_{i};s_{j}|C)}.\label{eq:REL_APROX_2}
\end{equation}
Notice that eq. (\ref{eq:REL_APROX_2}) is an approximation of the multidimensional MI expressed by eq. (\ref{eq:ed1}). Interestingly, Brown \textit{et al.} \cite{Brown2012} deduced a similar formula but with coefficients $\nicefrac{1}{|S^{t}|}$ replaced by unity constants.

Eq. (\ref{eq:REL_APROX_2}) allows deriving some well-known heuristic feature selection methods. When only the first two terms of Eq. (\ref{eq:REL_APROX_2}) are taken into account, it corresponds exactly to the minimal redundance maximal relevance (mRMR) criterion proposed in \cite{Peng05}. Moreover, if the term $\nicefrac{1}{|S|}$ is replaced by a user defined parameter $\beta$, then we obtain the MIFS criterion (\textit{Mutual Information Feature Selection}) proposed by Battiti \cite{Battiti94}. When considering only the first term in eq. (\ref{eq:REL_APROX_2}), we obtain the MIM criterion \cite{Lewis1992}.

Eq. (\ref{eq:REL_APROX_2}) with its three terms corresponds exactly to the Joint Mutual Information (JMI) \cite{YangMoody99,Brown2012}. Also it corresponds with the Conditional Infomax Feature Extraction (CIFE) criterion proposed in \cite{Lin2006}, when the coefficient $|S^{t}|=1, \forall t$. Moreover, the Conditional Mutual Information based Feature Selection (CMIFS) criterion proposed in \cite{Cheng2011} is an approximation of eq. (\ref{eq:REL_APROX_1}), where only 0, 1 or 2 out of $t$ summation terms are considered in each term. The CMIFS criterion is the following:
\begin{equation}
J_{cmifs}(f_{i})=I(f_{i};C)-I(f_{i};s_{t})+\sum_{s_{j}\in S;j\in\{1,t\}}I(f_{i};s_{j}|C)-I(f_{i};s_{t}|s_{1}). \label{cmifs}
\end{equation}

The previously mentioned methods do not take into account the terms containing $\neg s_{j}$ in eq. (\ref{eq:REL_APROX_1}). This entails the assumption that $f_{i}$ and $\neg s_{j}$ are independent, therefore ($I(f_{i};\neg s_{j})=I(f_{i};\neg s_{j}|C)=0$). This approximation can generate errors in the sequential selection or backward elimination of variables. In order to somehow take into account the missing terms, let us consider the following alternative approximation of $I(f_{i};C|S^{t})$:
\begin{eqnarray}
I(f_{i};C|S^{t}) & = & I(f_{i};C)+I(f_{i};S^{t};C)=\nonumber \\
 &  & I(f_{i};C)+I(f_{i};\{s_{j},\neg s_{j}\};C)=\nonumber \\
 &  & I(f_{i};C)+I(f_{i};s_{j};C)+I(f_{i};\neg s_{j};C|s_{j})=\nonumber \\
 &  & I(f_{i};C|s_{j})+I(f_{i};\neg s_{j};C|s_{j}).\label{eq:NO_LINEAL_SHANON}
\end{eqnarray}

Averaging this decomposition over every single feature $s_j \in S^{t}$ we have:
\begin{equation}
I(f_{i};C|S^{t})= \frac{1}{|S^{t}|}{\displaystyle \sum_{s_{j}\in S^{t}}I(f_{i};C|s_{j})}+\frac{1}{|S^{t}|}{\displaystyle \sum_{s_{j}\in S^{t}}I(f_{i};\neg s_{j};C|s_{j})}.\label{eq:CMIM}
\end{equation}

The Interaction Capping (ICAP) \cite{PHD_Jakulin2005} criterion approximates eq. (\ref{eq:NO_LINEAL_SHANON}) by the following expression:
\begin{equation}
J_{icap}(f_{i})=I(f_{i};C)+\sum_{s_{j}\in S}\min(0,I(f_{i};s_{j};C))).
\end{equation}

In ICAP \cite{PHD_Jakulin2005}, the information of variable $f_{i}$ is penalized when the interaction between $f_{i}$, $s_{j}$ and $C$ becomes redundant ($I(f_{i};s_{j};C)<0$), but the complementarity relationship among variables is neglected when $I(f_{i};s_{j};C)>0$. The authors considered a Naive Bayes classifier, which assumes independence between variables.

Eq. (\ref{eq:CMIM}) allows deriving the Conditional Mutual Information Maximization (CMIM) criterion \cite{Fleuret04}, when we consider only the first term on the right hand side of this equation and replace the mean operator with a minimum operator. CMIM discards the second term on the right hand side of eq.(\ref{eq:CMIM}) completely, taking into account only one-to-one relationships among variables and neglecting the multi-information among $f_{i}, \neg s_{j}$ and $C$ in the context of $s_j$ $\forall j$. On the other hand, CMIM-2 \cite{Vergara2010} criterion corresponds exactly to the first term on the right hand side of eq. (\ref{eq:CMIM}). These methods are able to detect pairs of relevant variables that act complementarily in predicting the class.  In general CMIM-2 outperformed CMIM in experiments using artificial and benchmark datasets \cite{Vergara2010}.

So far we have reviewed feature selection approaches that avoid estimating MI in high-dimensional spaces.
Bonev \textit{et al.} \cite{Bonev08} proposed an extension of the MD criterion, called  Max-min-Dependence (MmD), which is defined as follows:
\begin{equation}
J_{MmD}(f_{i})=I(\{f_{i},S\};C)-I(\neg \{f_{i},S\};C).
\end{equation}
The procedure starts with the empty set $S=\emptyset$ and sequentially generates $S^{t+1}$ as:
\begin{equation}
S^{t+1}=S^{t}\cup\underset{f_{i}\in F\backslash S}{\max}\left(J_{MmD}(f_{i})\right).
\end{equation}
The MmD criterion is heuristic, and is not derived from a principled approach. However, Bonev \textit{et al.} \cite{Bonev08} were one of the first in selecting variables estimating MI in high-dimensional spaces \cite{Hero99}, which allows using set of variables instead of individual variables. Chow and Huang \cite{Chow05} proposed combining a pruned Parzen window estimator with quadratic mutual information \cite{Principe}, using Renyi entropies, to estimate directly the MI between the feature subset $S^{t}$ and the classes $C$, $I(S^{t};C)$, in an effective and efficient way.

\section{Open Problems}
In this section we present some open problems and challenges in the field of feature selection, in particular from the point of view of information theoretic methods. Here can be found a non-exhaustive list of open problems or challenges.

\begin{enumerate}
\item \textbf{Further developing a unifying framework for information theoretic feature selection.}
As we reviewed in section \ref{UF}, a unifying framework able to explain the advantages and limitations of successful heuristics has been proposed. This theoretical framework should be further developed in order to derive new efficient feature selection algorithms that include in their functional terms information related to the three types of features: relevant, redundant and complementary. Also a stronger connection between this framework and the Markov blanket is needed. Developing hybrid methods that combine maximal dependency with minimal conditional mutual information is another possibility.

\item \textbf{Further improving the efficacy and efficiency of information theoretic feature selection methods in high-dimensional spaces.}
The computational time depends on the search strategy and the evaluation criterion \cite{GuyonFE}. As we enter the era of Big Data, there is an urgent need for developing very fast feature selection methods able to work with millions of features and billions of samples. An important challenge is developing more efficient methods for estimating MI in high-dimensional spaces. Automatically determining the optimal size of the feature subset is also of interest, many feature selection methods do not have a stop criterion. Developing new search strategies that go beyond greedy optimization is another interesting possibility.

\item \textbf{Further investigating the relationship between mutual information and Bayes error classification.}
So far lower and upper bounds for error classification have been found for the case of one random variable and the target class. Extending these results to the case of mutual information between feature subsets and the target class is an interesting open problem.

\item \textbf{Further investigating the effect of a finite sample over the statistical criteria employed and in MI estimation.}
Guyon \textit{et al.} \cite{GuyonFE} argued that feature subsets that are not sufficient may render better performance than sufficient feature subsets. For example, in the bio-informatics domain, it is common to have very large input dimensionality and small sample size \cite{Saeys07}.

\item \textbf{Further developing a framework for studying the relation between feature selection and causal discovery.}
Guyon \textit{et al.} \cite{GuyonCFS} investigated causal feature selection. The authors argued that the knowledge of causal relationships can benefit feature selection and viceversa. A challenge is to develop efficient Markov blanket induction algorithms for non-faithful distributions.

\item \textbf{Developing new criteria of statistical dependence beyond correlation and MI.}
Seth and Principe \cite{Seth2010} revised the postulates of measuring dependence according to Renyi, in the context of feature selection. An important topic is normalization, because a measure of dependence defined on different kinds of random variables should be comparable. There is no standard theory about MI normalization \cite{Estevez09,Duch2006}. Another problem is that estimators of measures of dependence should be good enough, even when using a few realizations, in the sense of following the desired properties of these measures. Seth and Principe \cite{Seth2010} argued that this property is not satisfied by MI estimators, because they do not reach the maximum value under strict dependence, and are not invariant to one-to-one transformations.
\end{enumerate}

\section{Conclusions}
We have presented a review of the state-of-the-art in information theoretic feature selection methods. We showed that modern feature selection methods must go beyond the concepts of relevance and redundance to include complementarity (synergy). In particular, new feature selection methods that assess features in context are necessary. Recently, a unifying framework has been proposed, which is able to retrofit successful heuristic criteria. In this work, we have further developed this framework, presenting some new results and derivations. The unifying theoretical framework allows us to indicate the approximations made by each method, and therefore their limitations.  A number of open problems in the field are suggested as challenges for the avid reader.

\section{Acknowledgement}
This work was funded by CONICYT-CHILE under grant FONDECYT 1110701.

\bibliographystyle{spbasic}
\bibliography{VergaraNCA}

\end{document}